\pdfoutput=1

\documentclass[11pt]{article}

\usepackage{emnlp2023}
\usepackage{times}
\usepackage{latexsym}

\usepackage[T1]{fontenc}

\usepackage{tikz}
\usepackage{tikz-dependency}
\usepackage[framemethod=tikz]{mdframed}
\usepackage{graphicx}
\usepackage{multirow}
\usepackage{arydshln}
\usepackage{lscape}
\usepackage{tipa}
\usepackage{transparent}
\usepackage{svg}

\usepackage{booktabs}

\usepackage[utf8]{inputenc}

\usepackage{csquotes}
\newcommand{\ia}{\textit{inter alia}}
\newcommand{\ie}{\textit{i.e.}}
\newcommand{\eg}{\textit{e.g.}}

\usepackage{microtype}

\usepackage{inconsolata}

\usepackage{enumitem}

\usepackage{etoolbox}
\apptocmd{\sloppy}{\hbadness 10000\relax}{}{}

\newcommand*{\role}[1]{\texttt{#1}}

\title{The Case for Scalable, Data-Driven Theory:\\A Paradigm for Scientific Progress in NLP}

\author{Julian Michael \\
  New York University \\
  \texttt{julianjm@nyu.edu} \\}

\begin{document}
\maketitle

\begin{abstract}
      I propose a paradigm for scientific progress in NLP
      centered around developing \textit{scalable, data-driven theories} of linguistic structure.
      The idea is to collect data in tightly scoped, carefully defined ways
      which allow for exhaustive annotation of behavioral phenomena of interest,
      and then use machine learning to construct explanatory theories
      of these phenomena which can form building blocks for intelligible AI systems.
      After laying some conceptual groundwork, I describe several investigations
      into data-driven theories of shallow semantic structure
      using Question-Answer driven Semantic Role Labeling (QA-SRL),
      a schema for annotating verbal predicate--argument relations using highly constrained
      question-answer pairs.
      While this only scratches the surface of the complex language behaviors of interest in AI,
      I outline principles for data collection and theoretical modeling
      which can inform future scientific progress.
      This note summarizes and draws heavily on my PhD thesis \citep{michael-2023-data}.
\end{abstract}

\section{Introduction}%

Formal representations of linguistic structure and meaning have long guided our understanding of how
to build NLP systems, \eg, in the traditional NLP pipeline \citep{jurafsky-martin-2008}.
However, this approach has always had limitations:
\begin{enumerate}[label=\arabic*.]
      \item Fully specifying formal representations requires resolving challenging theoretical questions
            long contentious among linguists;
      \item It is difficult to reliably produce these representations with broad coverage
            using machine learning; and,
      \item Even ostensibly correct linguistic representations are often hard to apply downstream.
\end{enumerate}
Together with the effectiveness of deep learning, these challenges
led to the proliferation of end-to-end neural network models which directly
perform tasks without intermediate formal representations of linguistic structure
\citep[][\ia]{he-etal-2017-deep,lee-etal-2017-end,seo-etal-2017-bidirectional}.
This trend continues with language model assistants like GPT-4 \citep{openai-2023-gpt4} and
Claude \citep{bai-etal-2022-constitutional} which can perform a wide range of tasks.
However, these systems are still not robust, often reporting false or biased answers
\citep{perez-etal-2022-discovering,bang-etal-2023-multitask}
and making false claims about their own reasoning \citep{turpin-etal-2023-language}.
Ensuring AI systems' robustness requires us to precisely characterize and control
their generalization behaviors.

To this end, formal theories, \eg, of linguistic structure, common sense, reasoning,
and world knowledge, provide frameworks for evaluation.
They inform the design and construction of challenge sets
\citep{mccoy-etal-2019-right,naik-etal-2018-stress,wang-etal-2019-glue}, measures of
systematicity \citep{yanaka-etal-2020-neural,kim-linzen-2020-cogs}, behavioral tests
\citep{linzen-etal-2016-assessing}, and probing experiments
\citep{liu-etal-2019-linguistic,tenney-etal-2019-what}.
As these theories allow us to characterize generalization behaviors we desire,
they will likely play a pivotal role in the design and training of trustworthy systems.
So core improvements in formal theories of aspects of intelligent behavior
may yield boons for both the construction and evaluation of NLP systems.
But the question remains of how to achieve this: decades of work on semantic
ontologies~\citep{baker-etal-1998-berkeley-framenet,palmer-etal-2005-proposition},
commonsense knowledge bases~\citep{lenat-1995-cyc,speer-etal-2017-conceptnet},
and formal reasoning systems~\citep{lifschitz-2008-what}
have largely been superseded in NLP by deep learning and language models.

Theory-driven approaches in AI have been so disappointing that \citet{sutton-2019-bitter} famously argues
that intelligence and the world are simply too complex for us to capture with domain theories,
and we should instead focus on general-purpose learning systems that can capture this intrinsic complexity from data.
However, I believe this is too pessimistic, giving up on the \textit{intelligibility}
of AI systems that is provided by accurate theories of their behavior,
which is necessary for verifying their safety and usefulness in high-risk,
high capability settings~\citep{ngo-etal-2023-alignment}.
Instead, the deep learning era presents an opportunity to rethink how we develop theories of language behavior.

In particular, I propose \textit{scalable, data-driven theory} as a paradigm
to address the shortcomings mentioned at the
beginning of this article: resolving or sidestepping theoretical questions,
producing representations with broad coverage,
and applying them effectively in downstream tasks.
Inspired by Pragmatist epistemology \citep{james-1907-pragmatism},
this approach avoids requiring the linguist or theoretician to specify the
entire theory by hand, instead integrating machine learning in a judicious way
which allows for the scalable, automated induction of formal theoretical constructs
(\eg, ontologies) which are grounded in task-relevant linguistic behaviors.

\section{Pragmatist Principles for Scientific Progress}%
\label{sec:pragmatism}

\citet{church-2007-pendulum} describes the history of computational linguistics
on a \textit{pendulum}, swinging between
Rationalist (theory-driven) and Empiricist (data-driven)
paradigms every 20 years.
\citeauthor{church-2007-pendulum} lists the ``swings'' as follows
      {\color{gray}(with my comments)}:
\begin{itemize}
      \item 1950s: Empiricism (Shannon, Skinner, Firth, Harris) {\color{gray}
                        --- information theory, psychological behaviorism, early corpus linguistics}
      \item 1970s: Rationalism (Chomsky, Minsky) {\color{gray}
                        --- generative linguistics, logic-based AI}
      \item 1990s: Empiricism (IBM Speech Group, AT\&T Bell Labs) {\color{gray}
                        --- statistical NLP, machine learning, modern distributional semantics}
      \item 2010s: A Return to Rationalism?
\end{itemize}
As the reader may know, the predicted ``Return to Rationalism'' did not happen.
NLP, for its part, is more Empiricist than ever.

Why is this?
\citeauthor{sutton-2019-bitter} may say it's because the world is too complex:
The Rationalist theoretician carefully formalizing the problems at hand
has no hope of capturing the world's intricacies in a manually-crafted theory,
though a system implementing that theory can be understood and controlled.
The Empiricist tinkerer, on the other hand, can build a system that mostly works
by trial, error, patching and fastening; so they win on empirical benchmarks.
However, the resulting system is too complex to fully understand or control,
and generalizes in unpredictable ways.

An odd feature of the Rationalism/Empiricism dichotomy is that
neither epistemology accurately describes the pursuit of science in most fields.
In fields like physics, chemistry, and biology, theoretical and experimental approaches are not in
conflict; rather, they synergize and inform each other, as theories are continually updated to align
with new experimental data.
To make sense of this, we can turn to an epistemology inspired by how people actually operate in
the world: Pragmatism.

\textit{Pragmatism} is an epistemological framework which conceptualizes
\textit{knowing} in terms of the \textit{actions} that the knowledge licenses, \ie, by the
predictions that follow from that knowledge. Prominent Pragmatists include Charles Sanders Peirce
(1839--1914) and William James (1842--1910).
Like Empiricism, Pragmatism embraces experience as the primary source of knowledge.
But unlike Empiricists, Pragmatists such as James embrace formal and linguistic
categories as comprising the content of knowledge, on the basis of their \textit{usefulness} in
making predictions and licensing actions \citep{james-1907-pragmatism}.
Unlike in Rationalism, the Pragmatist search for truth is not a search for
one true theory which fundamentally describes the world,
but for an ever-expanding set of theoretical tools and concepts that can be picked
up and put down according to the needs of the knower.
In pithy terms, a Pragmatist might agree with the statistical aphorism that that ``All models are
wrong; some are useful'' \citep{box-1976-science}.
Pragmatists such as \citet{james-1907-pragmatism} claim that this perspective more accurately
describes human behavior with respect to knowledge (and indeed, the pursuit of science) than prior
epistemologies.

Combining the core ideology of Pragmatism with observations from computational linguistics, we can
derive two guiding principles for the development of theories that may have prospective use in
NLP: decouple data from theory~(\autoref{sec:decouple}), and make data reflect use~(\autoref{sec:use}).

\subsection{Decouple Data from Theory}
\label{sec:decouple}

One feature that distinguishes much NLP work, particularly involving linguistic structure, from
traditional sciences is the status of theory with respect to data.
In most empirical sciences, data takes the form of concrete measurements of the world, and the task
of a theory is to explain those measurements.
In NLP, many benchmarks and datasets are constructed under the \textit{assumption} of a theory,
whether it be one of syntactic structure \citep{marcus-etal-1993-building,demarneffe2021universal},
semantic structure \citep{palmer-etal-2005-proposition,banarescu-etal-2013-abstract}, or some other
task-specific labeling scheme.

A theory, \eg, of syntactic or semantic structure, is useful for annotation in providing a
straightforward way to annotate disambiguation of text, which is important for understanding language.
However, errors and inconsistencies in annotation resulting from
complexity, vagueness, or underspecification in the theory limit what can be learned by
models, as human performance and inter-annotator agreement can be surprisingly low \citep{nangia-bowman-2019-human}.
For example, the OntoNotes compendium of semantic annotations \citep{hovy-etal-2006-ontonotes} was
presented as ``The 90\% solution'' because of 90\% agreement rates --- implying that the dataset
cannot validate performance numbers higher than 90\%.

As another example,
\citet{palmer-etal-2006-making} find that fine-grained sense distinctions produce
considerable disagreement among annotators of English text.
But fixing the problem can't just be a matter of improving the sense inventory:
they find that coarser-grained sense groups designed to improve agreement lack the distinctions from
fine-grained senses that are necessary for predicting how words should translate into
typologically distant languages like Chinese and Korean.
When different tasks require different theoretical distinctions, setting them in stone during
annotation is a problem,
especially considering that there will almost certainly be missing categories,
as new word senses or distinctions may show up in more exhaustive data or under domain shift.
More generally, refining annotation guidelines to increase agreement between annotators
does not necessarily solve the problem,
as the extra assumptions built into the annotation process do not necessarily
encode any more scientifically meaningful information in the data --- a problem
known in the philosophy of science as the
\textit{problem of theoretical terms}.\footnote{See \citet{riezler-2014-last} for a discussion of this issue
      in NLP.}

\newcommand{\delphin}{\textsc{delph-in}}

Building a robust theory that can scale to unexpected phenomena and new data, and be adjusted
for new tasks, requires theoretical agility which is precluded by committing to
a theory-based annotation standard.
An alternative is to directly annotate the phenomena that the theory is meant to explain, and derive
the theory on the basis of this data.
This, for example, is how \textit{grammar engineering} is done in the \delphin\ consortium
\citep{bender-emerson-2021-computational}.
For each language, a broad-coverage Head-driven Phrase Structure Grammar (HPSG) is maintained
separately from its associated treebank, which is annotated not with full syntactic analyses but
with \textit{discriminants} \citep{carter-1997-treebanker} such as prepositional phrase attachment
sites which constrain the set of possible parses in a way that is independent of the grammar.
Then, when the grammar is updated, the discriminants are used to automatically update the
treebank while also providing data to validate the updated theory
\citep{oepen-etal-2004-lingo,flickinger-etal-2017-sustainable}.
Pushing the envelope further are the Decompositional Semantics Initiative
\citep{white-etal-2016-universal} and MegaAttitude project
\citep{white-rawlins-2016-computational}.\footnote{\url{https://decomp.io}, \url{https://megaattitude.io}}
In these projects, annotating large-scale corpora with the phenomena that are posited to underly
linguistic theories in question --- such as \citet{dowty-1991-thematic}'s proto-role properties, or
entailments corresponding to neg-raising
\citep{an-white-2020-lexical} and projection \citep{white-rawlins-2018-role} ---
has facilitated insights regarding argument selection \citep{reisinger-etal-2015-semantic}
and lexically-specified
syntactic subcategorization rules \citep{white-2021-believing}, as well as automatically
inducing lexicon-level ontologies of semantic roles \citep{white-etal-2017-semantic} and event
structure \citep{gantt2021decomposing} that are derived directly from the phenomena they are
designed to explain.

The lesson of Empiricism is that for a model to work, it must be learned from data; while
Rationalism tells us that for a model to be intelligible and general, it must be grounded in theory.
A wealth of innovative prior work shows us that Pragmatism is possible: we can have both.

\subsection{Make Data Reflect Use}
\label{sec:use}

A satisfying data-driven theory of a few linguistic phenomena is not sufficient
as a backbone for general language understanding systems.
The second relevant lesson of Pragmatism is that the model must be fit to its use.
The approaches reviewed in \autoref{sec:decouple} are, by and
large, targeted at theoretical questions in language syntax and semantics,
\eg, regarding the nature of syntactic structure across many languages
\citep{bender-etal-2002-grammar} or the syntactic realization of a verb's arguments
\citep{reisinger-etal-2015-semantic}.
On the other hand, general-purpose language processing relies on a huge amount of lexical and world
knowledge and inferential ability which is outside the scope of traditional linguistic theories.
While general-purpose syntactic and semantic representations have some direct uses in NLP end-tasks,
such as for search and retrieval \citep{schafer-etal-2011-acl,shlain-etal-2020-syntactic}, their
application in downstream tasks requiring higher-level reasoning or inference, like reading
comprehension, translation, and information extraction has been less fruitful.
This is at least in part because these theories are far insufficient to serve as mechanistic
accounts of the inferential phenomena which are required to perform those
tasks.

Constructing theories which \textit{can} account for such phenomena is a monumental challenge. But
it is a challenge which, I argue, we must address if we want to pursue the goal of accurate,
reliable, and intelligible systems.
Pragmatism tells us the first step is to catalog the phenomena we wish to explain in a way
that is amenable to theoretical modeling.
This will require carefully carving up the space of phenomena in such a way that useful abstractions
can be designed to facilitate future progress~\citep{dijkstra-1974-role};
\autoref{sec:data} will discuss considerations on how to do this well.

\section{Scalable, Data-Driven Theory}
\label{sec:data-driven-theory}
The principles in \autoref{sec:pragmatism} imply a general framework for building useful theories,
which I call \textit{data-driven theory}:
First, annotate data in a theoretically-minimal way, scoped carefully to reflect specific phenomena that
we want to explain;
then, automatically induce theories to explain those phenomena using computational methods like machine learning.
But how does this method scale in practice?
Even if the resulting theories are high-quality, requiring annotated data
limits their scope to orders of magnitude less than what is leveraged by
standard pretrained
models~\citep{brown-etal-2020-language,openai-2023-gpt4,bai-etal-2022-constitutional}.

\paragraph{Black-Box Data Simulators}
This is where black-box models may actually be able to help.
Even if they are uninterpretable on their own,
their high accuracy and data efficiency means they can be used as
\textit{data simulators}, generating phenomenological data --- potentially
at a level of granularity or exhaustivity unobtainable from humans --- which can be fed
into another, more interpretable algorithm to distill a theory from it.
This is the approach we take in \citet{michael-zettlemoyer-2021-inducing},
described in \autoref{sec:theory}:
We first train a black-box model to generate QA-SRL questions,
where each role is labeled with only a single question in the training data.
Then we decode full question \textit{distributions} from this model,
and induce an ontology of semantic roles by clustering arguments based on the overlap of their
question distributions.
While this work required a large training set of QA-SRL annotations \citep{fitzgerald-etal-2018-large},
it may now be possible to do such experiments without large-scale human data annotation at all,
thanks to recent advances in instruction following by language models~\citep{openai-2023-gpt4,bai-etal-2022-constitutional}.

It may seem like the use of a black-box model as a data simulator begs the question:
if our concern is that the black-box model isn't learning the underlying function we hope it is,
then doesn't using it to simulate data risk leading us to a theory of the wrong function?
Well, yes --- \textit{but the theory lets us do something about it}.
Examining the ``wrong'' parts of the resulting theory (\eg, induced semantic roles
that don't match what we intuitively expect, or that lead to downstream predictions we think are wrong),
and their connection to the training data, will identify one of the following:
\begin{itemize}
      \item Systematic gaps in the data or mistakes in the model used for data simulation --- which can
            then be filled or corrected.
      \item Mistakes in the modeling assumptions used in the theory induction algorithm --- giving us
            information useful for improving our theories.
      \item Mistakes in our intuition about what the theory should have looked like in the first
            place --- which means we've learned something.
\end{itemize}
All of these are positive outcomes for scientific progress.
See \citet{michael-zettlemoyer-2021-inducing} for an in-depth analysis of this kind.

\paragraph{Scaling in Complexity}
Even if we can scale a theory's \textit{size}, \eg, to a large knowledge base or linguistic ontology,
this does not handle the case of more \textit{complex} tasks,
with more nuanced relations between input and output (such as open-ended question answering or
common sense inference tasks).
Since theoretical modeling requires narrowly-scoped data (discussed more in \autoref{sec:data}),
I do not expect that we can construct theories of such broad capabilities in the short term.
However, if we carve up the space of tasks to start with theories of simple sub-phenomena
of reading and inference,
then we may be able to bootstrap from these theories to annotate and make sense of more
complex data --- for example,
one can imagine eventually inducing rich, broad-coverage entailment graphs in the style
of \citet{berant-etal-2015-efficient} or \citet{mckenna-etal-2023-smoothing}
on the basis of comprehensive annotations of structured inferences in context.
A complete or ``true'' theory of complex NLP tasks may be impossible even in
principle, but --- in the spirit of Pragmatism --- that doesn't mean we can't construct theories
that are \textit{useful} for understanding and controlling AI systems.
How my proposed framework scales with task complexity is unclear as of yet,
but scalable theories of narrow phenomena provide a step in the right direction.

\section{Data: Scoping Language Behaviors}%
\label{sec:data}
The first step to developing theories of linguistic structure in an empirical, data-driven way
is to carefully choose the data.
To guide this, I propose \textbf{Four Principles of Scientific Data for NLP}:
\begin{enumerate}
      \item \textbf{Theoretical minimalism.} The data should rely on as few theoretical assumptions as
            possible. For example, to capture natural language syntax, you should directly annotate
            the \textit{phenomena} that you intend your syntactic theory to explain rather than
            directly annotating theoretical constructs like syntactic trees.
            This creates the space for an underlying theory to meaningfully explain this data.
      \item \textbf{Broad comprehensibility.} To facilitate on-demand data collection at large scale
            in new domains, it should be possible and affordable to recruit non-expert annotators
            to label large amounts of data (\eg, through crowdsourcing),
            or it should be feasible to automatically generate the data (\eg, with language models).
      \item \textbf{Annotation constraints.} The output space of the task should be sufficiently constrained
            to allow for exhaustive coverage of the phenomena of interest.
            A task which is too open-ended
            leads annotators to produce a convenience sample of the output space,
            resulting in biased data that doesn't capture the full complexity of the phenomena of
            interest~\citep{cai-etal-2017-pay,gururangan-etal-2018-annotation}.
      \item \textbf{Narrow scope.}
            The task should not capture too much complexity in the relationship between input and output.
            Not only can this make it difficult for annotators to reliably produce high-quality data,
            but it makes it more difficult to model the phenomena expressed in the data with a comprehensible
            theory.
\end{enumerate}

Principles 1 and 2 instantiate \autoref{sec:decouple}'s recommendation to decouple data from theory,
while Principles 2, 3 and 4 help make it tractable to develop broad-coverage, comprehensible theories from this data.
The final requirement is that the data reflect relevant downstream use cases (\autoref{sec:use}),
which in our case means it should encode phenomena representing the intended behavior of AI systems
performing language tasks.\footnote{This work is concerned with normative theories of AI behavior when
      performing language tasks. Insofar as we wish to produce theories of AI behavior which are comprehensible to us,
      aligned with our intuitions, and allow us to interface fluidly with machines using language,
      this goal should mostly be aligned with developing \textit{descriptive} theories of \textit{human} language behavior,
      which can then be used to constrain and guide AI behavior. The relationship between these theories and
      their importance for interacting with machines are discussed more in Chapter 2 of \citet{michael-2023-data}.}
I focus on a key strategy to meet these requirements:
\textit{annotating natural language with natural language question-answer pairs}.
Question answering has long been used as a general-purpose format for
testing language comprehension or executing practical language
tasks \citep{gardner-etal-2019-question,mccann-etal-2018-natural,mccarthy-1976-example},
as nearly any task can be phrased as a question and questions which test a reader's comprehension of a text
need not require specialized linguistic or theoretical expertise to answer.
The downside of this great generality is that data annotation tends to be
highly under-constrained and unsystematic \citep{gardner-etal-2019-making},
so we must judiciously constrain the space of question-answer pairs we use
in accordance with the Four Principles.

This work is focused on annotations of shallow semantic structure:
syntax, semantic roles, and other predicate--argument structure relations
expressed in text.
\citet{he-etal-2015-question} pioneered the use of question-answer pairs as a proxy for such structure
in \textit{Question-Answer driven Semantic Role Labeling} (QA-SRL),
a framework for annotating English verbal predicate--argument relations using simple, highly
constrained question-answer pairs.
In the rest of this section, I will describe three data annotation projects which
explored variations of this approach,
illustrating some of the basic tensions between the Four Principles.

\def\subfigspace{\vspace{-1em}}

\subsection{Human-in-the-Loop Parsing}

\begin{figure}
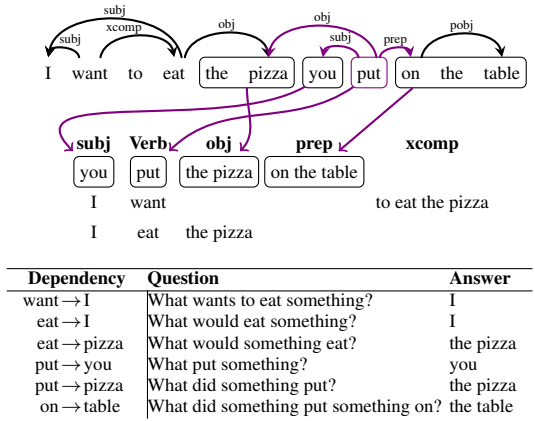

      \centering
      \small
      \begin{dependency}[theme=simple,edge style=thick,font=\scriptsize]
            \begin{deptext}[column sep=0.4em]
                  I \& want \& to \& eat \& the \& pizza \& you \& put \& on \& the \& table \\
            \end{deptext}
            \depedge[label style={shift={(0.1em,0em)}}]{2}{1}{subj} 
            \depedge[label style={shift={(-0.2em,0em)}}]{2}{4}{xcomp}
            \depedge[edge start x offset=.75em,edge unit distance=3ex]{4}{1}{subj} 
            \depedge{4}{6}{obj}
            \depedge[edge start x offset=.75em,violet]{8}{6}{obj} 
            \depedge[violet]{8}{7}{subj}
            \depedge[violet]{8}{9}{prep}
            \depedge{9}{11}{pobj}

            \wordgroup{1}{7}{7}{parse-you}
            \wordgroup[violet]{1}{8}{8}{parse-put}
            \wordgroup{1}{5}{6}{parse-thepizza}
            \wordgroup{1}{9}{11}{parse-onthetable}

            \begin{deptext}[column sep=0.3em,row sep=0em,below of=\matrixref]
                  \\[2em]
                  \textbf{subj} \& \textbf{Verb} \&
                  \textbf{obj} \& \textbf{prep} \& \textbf{xcomp} \\
                  you \& put \& the pizza \& on the table \& \\
                  I \& want \& \& \& to eat the pizza \\
                  I \& eat \& the pizza \& \& \\
            \end{deptext}
            \wordgroup{3}{1}{1}{phrase-you}
            \wordgroup{3}{2}{2}{phrase-put}
            \wordgroup{3}{3}{3}{phrase-thepizza}
            \wordgroup{3}{4}{4}{phrase-onthetable}

            \draw [->, thick, violet] (parse-you){}+(-0.75em,-0.6em) to[in=135,out=-150] (phrase-you);
            \draw [->, thick, violet] (parse-put){}+(-0.65em,-0.6em) to[in=45,out=-145] (phrase-put);
            \draw [->, thick, violet] (parse-thepizza){}+(0,-0.65em) to[in=45,out=-90] (phrase-thepizza);
            \draw [->, thick, violet] (parse-onthetable){}+(-2em,-0.65em) -- (phrase-onthetable);

      \end{dependency}



      \vspace{0.5em}
      {\scriptsize
            \begin{tabular}{r@{\hskip 0.1em}c@{\hskip 0.1em}l|@{\hskip 0.4em}l@{\hskip 0.4em}l}
                  \hline
                  \multicolumn{3}{c}{\textbf{Dependency}} & \textbf{Question}                    & \textbf{Answer} \\ \hline
                  want                                    & \( \to \)                            & I
                                                          & What wants to eat something?         & I               \\
                  eat                                     & \( \to \)                            & I
                                                          & What would eat something?            & I               \\
                  eat                                     & \( \to \)                            & pizza
                                                          & What would something eat?            & the pizza       \\
                  put                                     & \( \to \)                            & you
                                                          & What put something?                  & you             \\
                  put                                     & \( \to \)                            & pizza
                                                          & What did something put?              & the pizza       \\
                  on                                      & \( \to \)                            & table
                                                          & What did something put something on?
                                                          & the table                                              \\
            \end{tabular}}

      \vspace{1em}

      \caption{
            Question-answer pair generation for human-in-the-loop parsing \citep{he-etal-2016-human}.
            We use the predicted CCG category of each verb to generate the questions,
            which are in in one-to-one relation with syntactic dependencies in the sentence.
            This one-to-one assumption was ultimately too strong, as workers answer these questions according
            to semantics and not just syntax.}

      \subfigspace

      \label{fig:hitl-qa-gen-example}
\end{figure}

\citet{he-etal-2016-human} introduces \textit{human-in-the-loop parsing}.
We construct multiple-choice questions from syntactic attachment ambiguities in a parser's $n$-best list,
get crowdsourced workers to answer these questions,
and then re-parse the original sentence with constraints derived from the results (\autoref{fig:hitl-qa-gen-example}).
Testing on the English CCGbank~\citep{hockenmaier-steedman-2007-ccgbank},
we find only a small improvement in parser performance.
A core challenge is the \textit{syntax--semantics mismatch},
where workers provide answers which are semantically correct but
correspond to the wrong syntactic attachment.
For example, in the sentence
``Kalipharma is a New Jersey--based pharmaceuticals concern that sells products under the Purepac label'',
workers unanimously answer the question ``What sells something?'' with ``Kalipharma'',
which is not the syntactic subject of \textit{sells} but a more natural way of referring to the same entity.
So even though our annotation task is tightly scoped,
our interpretation of the results requires theoretical assumptions
which do not match the intuitions of non-expert workers.

\subsection{Crowdsourcing Question-Answer Meaning Representations}

\newmdenv[innerlinewidth=0.5pt, roundcorner=2pt,linecolor=black,innerleftmargin=6pt,
      innerrightmargin=6pt,innertopmargin=6pt,innerbottommargin=6pt]{examplebox}

\begin{figure}[t!]
      \small
      \centering
      \begin{examplebox}
            Pierre Vinken, 61 years old, will join the board as a nonexecutive director Nov. 29.
            \vspace{5pt}
            \textcolor{black}{\hrule height 0.6pt}
            \vspace{5pt}
            Who will \textbf{join} as \textbf{nonexecutive director}? - Pierre Vinken \\
            What is \textbf{Pierre}'s last name? - Vinken \\
            Who is \textbf{61 years old}? - Pierre Vinken \\
            How \textbf{old} is \textbf{Pierre Vinken}? - 61 years old \\
            What will he \textbf{join}? - the board \\
            What will he \textbf{join the board} as? - nonexecutive director \\
            What type of \textbf{director} will \textbf{Vinken} be? - nonexecutive \\
            What day will \textbf{Vinken join the board}? - Nov. 29
      \end{examplebox}
      \caption{Example Question-Answer Meaning Representation \citep{michael-etal-2018-crowdsourcing}.
            Non-stopwords drawn from the source sentence are in bold. QAMR question--answer pairs capture
            a wide variety of relations, but are unstructured and hard to use downstream without extra tools
            such as a syntactic parser --- here, our annotation task was too unconstrained and task scope too broad.}
      \label{fig:qamr-example}

      \subfigspace
\end{figure}

\citet{michael-etal-2018-crowdsourcing} takes the opposite tack,
broadening the task's scope by gathering open-ended questions from annotators to capture
as many semantic relationships as possible in the source sentence.
This requires adding many careful constraints and incentives to the
crowdsourcing procedure, but we are careful to
allow for open-ended questions that express annotator creativity.
The result is a dataset of \textit{Question-Answer Meaning Representation} (QAMR) annotations
over English encyclopedic and news text covering many interesting phenomena (see \autoref{fig:qamr-example}).
However, achieving high recall of predicate--argument relations is not economical,
requiring high annotation redundancy,
and the unstructured question-answer pairs are hard to use downstream.
The most successful use of QAMR in follow-up work
is probably \citet{stanovsky-etal-2018-supervised},
where we convert QAMRs into Open Information Extraction tuples, but have to run
the questions through a syntactic parser to do so.
The lesson from these results is that leaving the annotation space too open and unconstrained
leads to difficulties with recall and challenges with downstream modeling and theory.

\subsection{Large-Scale QA-SRL Parsing}
\label{sec:qasrl}

\begin{table*}
      \centering
      \small
      \textit{The plane was \textbf{diverting} around weather formations over the Java Sea when \\ contact
            with air traffic control (ATC) in Jakarta was \textbf{lost}.} \\[0.5em]
      \begin{tabular}{@{}rrrrrrrrl@{}}
            \toprule
            \textbf{wh} & \textbf{aux} & \textbf{subj} & \textbf{verb}
                        & \textbf{obj} & \textbf{prep} & \textbf{obj2}  & \textbf{?} & \textbf{Answer}                                                    \\
            \midrule
            What        & was          &               & being diverted &            & around          &  & ? & \textit{weather formations}               \\
            What        & was          &               & diverting      &            &                 &  & ? & \textit{The plane}                        \\
            What        & was          &               & being diverted &            &                 &  & ? & \textit{The plane}                        \\
            What        & was          &               & lost           &            &                 &  & ? & \textit{contact with air traffic control} \\
            Where       & was          & something     & lost           &            &                 &  & ? & \textit{over the Java Sea}                \\
            \bottomrule
      \end{tabular}
      \caption{QA-SRL question-answer pairs from the development set of the QA-SRL Bank 2.0 \citep{fitzgerald-etal-2018-large}.
            We constrained the questions with a non-deterministic finite automaton (NFA) encoding English clause structure
            for question autocomplete and auto-suggest.
            This facilitated high-quality, high-coverage annotation at scale
            while providing the expressiveness to represent the semantic role relations within each sentence.}
      \label{tab:qasrl-format}
      \subfigspace
\end{table*}

\citet{fitzgerald-etal-2018-large} returns to QA-SRL.
In the original QA-SRL work~\citep{he-etal-2015-question},
trained annotators specify the questions using drop-down menus in an excel spreadsheet.
In this work, we streamline and scale up data collection,
gathering high-coverage annotations for over 64,000 sentences
with a two-stage generate/validate crowdsourcing pipeline (see \autoref{tab:qasrl-format} for examples).
We increase annotation speed, reliability, and coverage using
an autocomplete system which tracks the syntactic
structure of QA-SRL questions as the annotator types,
using it to suggest completions as well as whole questions.
In terms of semantic richness and annotation constraints,
these annotations are somewhere between
our work on human-in-the-loop parsing and question-answer meaning representations.
The constrained task and high coverage allow us to train high-quality QA-SRL predictors
and enables future work
on semantic role induction~(\autoref{sec:role-induction})
and controlled question generation~(\autoref{sec:role-questions}).

\paragraph{Takeaways} 
Our results over the course of these projects suggests that we should search for tasks
in a ``goldilocks zone'':
Their scope should not be so constrained or beholden to prior theory as to be unintuitive,
but not so unconstrained that it is hard to get exhaustive and reliable annotation of interesting
phenomena.
As annotation constraints depend on \textit{some} prior theory of the phenomena to be captured,
these constraints need to be carefully chosen so as to minimize arbitrary assumptions
in the task setup and make sure the task is natural for annotators.
In the case of QA-SRL, the prior theory we incorporated is a small grammar fragment of English
encompassing QA-SRL questions.
Our findings support that QA-SRL, with the annotation aids
developed in \citet{fitzgerald-etal-2018-large},
strikes a good balance of the Four Principles.

\section{Theory: From Language, Structure}%
\label{sec:theory}

In this section, I will describe two projects
which show how QA-SRL can be used to build a data-driven theory
which is directly applicable in downstream tasks.

\subsection{Inducing Semantic Roles Without Syntax}
\label{sec:role-induction}

\begin{table}[t]
      \small
      \centering
      \begin{tabular}{@{}llr@{}}
            \toprule
            \textbf{Labels}    & \multicolumn{2}{l}{\hspace{-.5em}\textbf{Questions}}                    \\ 
            \midrule
            \role{A1} (98\%) 
                               & \hspace{-.5em}What is given?                                      & .30 \\
                               & \hspace{-.5em}What does something give something?\hspace{-1em}    & .21 \\
                               & \hspace{-.5em}What does something give?                           & .20 \\
                               & \hspace{-.5em}What is something given?                            & .11 \\
            \cmidrule{2-3}
            \role{A0} (98\%)
                               & \hspace{-.5em}What gives something?                               & .44 \\
                               & \hspace{-.5em}What gives something something?                     & .27 \\
                               & \hspace{-.5em}What gives something to something?                  & .08 \\
            \cmidrule{2-3}
            \role{A2} (94\%)
                               & \hspace{-.5em}What is given something?                            & .28 \\
                               & \hspace{-.5em}What does something give something to?\hspace{-1em} & .18 \\
                               & \hspace{-.5em}What does something give something?                 & .14 \\
                               & \hspace{-.5em}What is given?                                      & .09 \\
                               & \hspace{-.5em}What is something given to?                         & .07 \\
            \cmidrule{2-3}
            \role{TMP} (46\%),
                               & \hspace{-.5em}When does something give something?                 & .20 \\
            \role{ADV} (22\%), & \hspace{-.5em}How does something give something?                  & .09 \\
            \role{MNR} (12\%)  & \hspace{-.5em}When is something given?                            & .09 \\
                               & \hspace{-.5em}When is something given something?                  & .09 \\
            \cmidrule{2-3}
            \role{PNC} (30\%),%
                               & \hspace{-.5em}Why does something give something?                  & .18 \\
            \role{ADV} (22\%), & \hspace{-.5em}Why does something give up something?\hspace{-1em}  & .07 \\
            \role{TMP} (14\%)  & \hspace{-.5em}Why is something given something?                   & .07 \\
            \bottomrule
      \end{tabular}
      \caption{Roles for \textit{give} produced by \citet{michael-zettlemoyer-2021-inducing}.
            For each predicate, we cluster its arguments in PropBank
            based on the similarity of the distributions of QA-SRL questions our model generates.
            In this case, core arguments are captured almost perfectly, exhibiting both passive and dative alternations.
      }
      \label{tab:give-roleset}
      \subfigspace
\end{table}

\citet{michael-zettlemoyer-2021-inducing}
show how to use QA-SRL to automatically induce an ontology of semantic roles,
leveraging a key insight: the \textit{set} of QA-SRL questions that are
correctly answered by a given answer span identifies an underlying semantic role
through its syntactic alternations, which are representative of the phenomena that
a semantic role ontology like PropBank is designed to explain.
We leverage this insight by using a trained QA-SRL question generator as a data
simulator, generating a full distribution over (simplified)
QA-SRL questions for each argument of a verb appearing through an entire corpus.
Clustering these distributions of questions according to a simple maximum-likelihood objective
yields a set of discrete semantic roles that exhibits high agreement with existing resources
(see \autoref{tab:give-roleset}).
This presents an approach which could potentially be used to develop semantic role ontologies
in new domains where they are not currently available,
with directions for improving QA-SRL data
toward the end of automatically inducing better semantic roles.

\begin{figure*}[t]
      \centering
      \includesvg[width=0.9\textwidth]{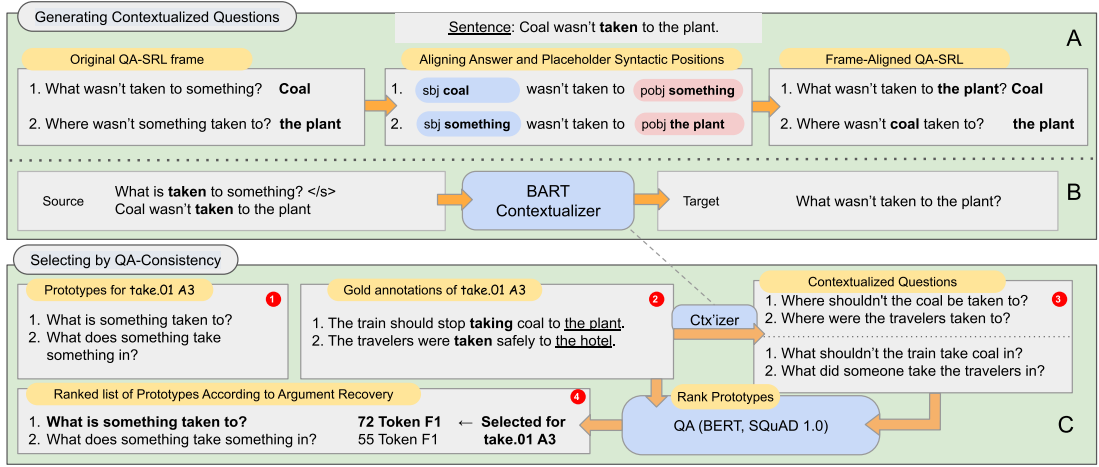}
      \caption{
            Overview of \citet{pyatkin-etal-2021-asking}'s approach.
            The natural correspondence between QA-SRL questions and semantic roles allows us to use QA-SRL question templates
            in a planning step to successfully generate questions for any PropBank semantic role,
            even when the corresponding argument doesn't appear in the source sentence
            (a situation never encountered in training data).
            \textbf{A: Construction of Frame-Aligned QA-SRL}
            using syntactic information inferred by the autocomplete NFA from \citet{fitzgerald-etal-2018-large}, \ie,
            leveraging our (minimal) theoretical assumptions about argument structure.
            \textbf{B: Contextualizing questions} by feeding a prototype question and context into a neural model
            that outputs a Frame-Aligned QA-SRL question.
            \textbf{C: Selecting prototype questions} by testing each prototype (1) against a sample of arguments for each role (2).
            After contextualization (3), each question is fed into a QA model and we choose the prototype that most often recovers the
            correct argument (4).}
      \label{fig:roleqs-overview}
      \subfigspace
\end{figure*}

\subsection{Asking it All: Generating Contextualized Questions for any Semantic Role}
\label{sec:role-questions}

\citet{pyatkin-etal-2021-asking} use QA-SRL to build a controllable question generation system.
The task is to generate fluent questions asking about the arguments corresponding to specific
semantic roles in context (see \autoref{fig:roleqs-overview} for an overview).
The challenge is a lack of training data, as QA-SRL
questions are not fully natural
and are not annotated for roles which aren't expressed in a sentence.
We leverage two key insights:
First, we find that QA-SRL questions generally correspond to the same role across many contexts.
So we prime our question generation system with a template QA-SRL question
corresponding to the correct role, leading it to generate semantically correct questions
even when the answer isn't present in the sentence. 
Second, we use the syntactic structure of QA-SRL questions to
align the placeholders (\textit{someone}, \textit{something})
in each question with the answers of other questions,
translating QA-SRL questions into more fluent ones closer to those in QAMR.

\paragraph{Takeaways}
Together this work illustrates not only the promise for the development of large-scale ontologies
in a data-driven way~(\autoref{sec:role-induction}),
but it also illustrates how having these ontologies computationally grounded in the phenomena they
are designed to explain, \ie, question-answer pairs,
facilitates ontology's the downstream use~(\autoref{sec:role-questions}).
It's not hard to imagine next steps incorporating an induced ontology of semantic roles
into \citet{pyatkin-etal-2021-asking}'s system
to obviate the need for a pre-specified role ontology altogether.

\section{Concluding Thoughts}%
\label{sec:conclusion}

I have proposed \textit{scalable, data-driven theory} as a Pragmatist paradigm for
scientific progress in NLP.
To develop scalable theories, one should:
\begin{enumerate}[label=\arabic*.]
      \item Collect carefully-scoped data that directly represents a phenomenon of interest
            while imposing minimal prior theoretical assumptions,
      \item Increase the data's scale and coverage using a learned black-box data simulator,
      \item Induce comprehensible models of this high-coverage data with machine learning, and
      \item Examine the results to debug and improve the theory and data,
            progressing our scientific understanding of the phenomenon of interest.
\end{enumerate}
Using QA-SRL, I have shown how to leverage black-box data simulation together with
simple probabilistic modeling to automatically induce an ontology of semantic roles
which is directly and comprehensibly grounded in phenomena that the theory of semantic roles
is meant to explain.
This not only lays the groundwork for new scalable theoretical developments in semantic
representation,
but can serve as an example to guide future work on scalable theories in other
domains.

\subsection*{Why now?}
The justification for building scalable, data-driven theories can be
summarized as follows:
\begin{enumerate}
      \item To build systems which generalize in controllable, predictable ways,
            we need comprehensible theories of their desired behavior.
      \item However, the behaviors we wish to produce in AI and NLP are too
            complex for us to easily write down theories of how they should work.
      \item So instead, we must use machines (\ie, statistical models) to
            construct our theories on the basis of data in a scalable way.
            The role for the scientist here is twofold:
            \begin{itemize}
                  \item to carefully determine the scope of the phenomena to be
                        explained and curate the data accordingly, and
                  \item to define the meta-theory which relates the learned theory
                        to the data.
            \end{itemize}
\end{enumerate}
This argument could have been made at any point in the history of NLP,
so why do I make it now?\footnote{
      Similar arguments have been made before in grammar engineering
      \citep{oepen-etal-2004-lingo,flickinger-etal-2017-sustainable}
      and the Decompositional Semantics Initiative \citep{white-etal-2016-universal},
      while in linguistic typology, \citet{haspelmath-2010-framework}'s
      \textit{framework-free grammatical theory}
      makes similar points about the relationship between data and theory.
      My approach differs from these in my focus on applications in NLP
      where the vastness and complexity of the domain
      becomes more of a challenge.}
I think the argument would have been viewed as premature
in the \textit{era of underfitting} prior to the deep learning revolution.
Statistical models like CRFs \citep{lafferty-etal-2001-conditional}
struggle even in-distribution on tasks like syntactic and semantic parsing,
let alone complex end tasks involving question answering or language generation.
The problem at that time was to build models expressive enough to perform well
while tractable enough to learn from data.
Pre-neural systems were weak enough that many thought they would benefit from
hand-curated linguistic resources like PropBank~\citep{palmer-etal-2005-proposition}.

With deep learning, these factors all changed:
the limits of hand-curated resources like PropBank have been surpassed,
and neural models fit all kinds of data distributions,
leaving us face-to-face with the problem of generalization
and the need for data-driven theory.
Furthermore, we have new tools for data simulation;
the role induction algorithm in \citet{michael-zettlemoyer-2021-inducing}
would not have been workable without a neural model to simulate dense
annotation of QA-SRL questions.
So we are finally in a position to make such theories scalable.

\subsection*{Looking forward}

As argued above, a critical role for the scientist in developing data-driven theories is to
define scopes of phenomena to be explained, carving linguistic behavior at useful joints.
I hope to have demonstrated that
the concept of \textit{semantic roles} provides such a useful scope,
where its corresponding phenomena (as QA-SRL)
can be effectively annotated at scale (\autoref{sec:qasrl}),
tractably modeled with a comprehensible theory (\autoref{sec:role-induction}),
and used for downstream tasks (\autoref{sec:role-questions}).
Moving forward requires carefully choosing more such useful concepts
and using them to scope phenomena, define and induce theories,
and tie these data and theories into downstream applications.

Extending the paradigm of scalable theory to more facilities of language (\eg,
syntax, word sense, or coreference)
and more complex phenomena (\eg, representations of
world knowledge, common sense, or reasoning)
remains a major challenge.
As the scope of the phenomena to be represented increases,
greater annotation constraints will be necessary in order to ensure that
these phenomena are adequately covered.
However, doing so while maintaining theoretical minimalism is challenging.
My hope is that scalable theories of narrowly-scoped subphenomena
(\eg, semantic roles) will provide constraints that make more complex
tasks tractable to exhaustively annotate, without introducing the
same problems as in the Rationalist paradigm where inconsistencies,
underspecification, and arbitrary theoretical choices limit the usefulness of the data.
In this way, it may be possible to bootstrap from narrowly-scoped theories
into progressively broad accounts of language structure, meaning, and intelligent behavior.

At this point, such talk is speculation.
It is unclear how data-driven theory will generalize to more complex tasks.
However, in this work I hope to have provided an argument this kind of work
is at least worth attempting,
and perhaps laid some groundwork and principles which can be used as a starting point for
it to be done in the future.

\section*{Acknowledgments}
Thanks to my PhD thesis advisor Luke Zettlemoyer, as well as
reading committee members Noah A. Smith and Emily M. Bender,
and committee member Shane Steinert-Threlkeld.
Many thanks also to my collaborators on the projects reviewed
in this note, including Ido Dagan, Luheng He, Gabriel Stanovsky,
Valentina Pyatkin, Paul Roit, and Nicholas FitzGerald,
and others who have done essential QA-Sem work following on QA-SRL,
including Ayal Klein and Daniela Weiss,
as well as the many annotators who have contributed to
building these datasets.
Thanks also to my brother Jonathan Michael for introducing me to Pragmatism
and Ari Holtzman for helpful and engaging discussions about it.
Finally, thanks to the anonymous reviewers for helpful comments on
what I should include in this note to round out the discussion.
See \citet{michael-2023-data} for more detailed acknowledgments
for my thesis work.

\bibliographystyle{acl_natbib}
\bibliography{anthology,references}

\end{document}